\documentclass[runningheads]{llncs}

\usepackage{graphicx}

\usepackage{booktabs}
\usepackage{todonotes}
\usepackage{comment}    
\usepackage{algpseudocode}
\usepackage{multirow}
\usepackage{mathtools}
\usepackage{algpseudocode}
\usepackage{graphicx}
\usepackage{subfig}
\usepackage{tabulary}

\usepackage{amsmath,amssymb,trimclip,adjustbox}

\usepackage{tikz}
\usetikzlibrary{automata, positioning}
\usetikzlibrary{shapes.geometric, arrows}
\usepackage{textcomp}
\usepackage{pgfplots}

\usepackage[linesnumbered,ruled,vlined]{algorithm2e}

\SetCommentSty{mycommfont}
\SetKwRepeat{Do}{do}{while}
\SetKwInput{KwInput}{Input}                
\SetKwInput{KwOutput}{Output}              

\tikzstyle{startstop} = [rectangle, rounded corners, minimum width=1cm, minimum height=0.5cm,text centered, draw=black]
\tikzstyle{io} = [trapezium, trapezium left angle=70, trapezium right angle=110, minimum width=2cm, minimum height=0.5cm, text centered, draw=black]
\tikzstyle{process} = [rectangle, minimum width=2cm, minimum height=0.5cm, text centered,draw=black]
\tikzstyle{process_n} = [rectangle, minimum width=1.5cm, minimum height=0.5cm, text centered,text width=1.5cm, draw=black]
\tikzstyle{decision} = [diamond, minimum width=1.2cm, minimum height=0.5cm, text centered, aspect=1.5, draw=black]
\tikzstyle{arrow} = [->,>=stealth]
\def\BibTeX{{\rm B\kern-.05em{\sc i\kern-.025em b}\kern-.08emT\kern-.1667em\lower.7ex\hbox{E}\kern-.125emX}}

\begin{document}
\title{Probabilistic Verification of Neural Networks Against Group Fairness}

\author{Bing Sun\inst{1}, Jun Sun\inst{1}, Ting Dai\inst{2}, and Lijun Zhang\inst{3}}
\institute{$^1$Singapore Management University \\
$^2$Huawei Singapore\\
$^3$Chinese Academy of Science}

%
%
%
%
%
\maketitle              
\begin{abstract}
Fairness is crucial for neural networks which are used in applications with important societal implication. Recently, there have been multiple attempts on improving fairness of neural networks, with a focus on fairness testing (e.g., generating individual discriminatory instances) and fairness training (e.g., enhancing fairness through augmented training). In this work, we propose an approach to formally verify neural networks against fairness, with a focus on independence-based fairness such as group fairness. Our method is built upon an approach for learning Markov Chains from a user-provided neural network (i.e., a feed-forward neural network or a recurrent neural network) which is guaranteed to facilitate sound analysis. The learned Markov Chain not only allows us to verify (with Probably Approximate Correctness guarantee) whether the neural network is fair or not, but also facilities sensitivity analysis which helps to understand why fairness is violated. We demonstrate that with our analysis results, the neural weights can be optimized to improve fairness. Our approach has been evaluated with multiple models  trained on benchmark datasets and the experiment results show that our approach is effective and efficient.
\end{abstract}
\section{Introduction}
In recent years, neural network based machine learning has found its way into various aspects of people's daily life, such as fraud detection~\cite{fraud_detection}, facial recognition~\cite{face_recognition}, self-driving~\cite{selfdriving}, and medical diagnosis~\cite{medical_diagnosis}. Although neural networks have demonstrated astonishing performance in many applications, there are still concerns on their dependability. One desirable property of neural networks for applications with societal impact is fairness~\cite{trust_ai}. Since there are often societal biases in the training data, the resultant neural networks might be discriminative as well. This has been demonstrated in~\cite{fairtest}. Fairness issues in neural networks are often more `hidden' than those of traditional decision-making software programs since it is still an open problem on how to interpret neural networks.

Recently, researchers have established multiple formalization of fairness regarding different sub-populations~\cite{demographic,verification,fairness,counterfactual}. These sub-populations are often determined by different values of protected features (e.g., race, religion and ethnic group), which are application-dependent. To name a few, group fairness requires that minority members should be classified at an approximately same rate as the majority members~\cite{demographic,verification}, whereas individual discrimination (a.k.a.~causal fairness) states that a machine learning model must output approximately the same predictions for instances which are the same except for certain protected features~\cite{fairness,counterfactual}. We refer readers to~\cite{science} for detailed definitions of fairness. In this work, we focus on an important class of fairness called independence-based fairness, which includes the above-mentioned group fairness.

Recently, there have been multiple attempts on analyzing and improving fairness of neural networks, with a focus on fairness testing (e.g., generating individual discriminatory instances) and fairness training (e.g., enhancing fairness through augmented training). Multiple attempts~\cite{THEMIS,aequitas2018,SG,adf2020} have been made on testing machine learning models against individual discrimination, which aims to systematically generate instances that demonstrate individual discrimination. While these approaches have impressive performance in terms of generating such instances, they are incapable of verifying fairness. Another line of approaches is on fairness training~\cite{ftrain1,ftrain2,ftrain3,ftrain4,embedding,counterfactual}, this includes approaches which incorporate fairness as an objective in the model training phase~\cite{ftrain1,ftrain2,ftrain3}, and approaches which adopt heuristics for learning fair classifiers~\cite{ftrain4}. While the experiment results show that these approaches improve fairness to certain extent, they do not guarantee that the resultant neural networks are fair.  

In this work, we investigate the problem of verifying neural networks against independence-based fairness. Our aim is to design an approach which allows us to (1) show evidence that a neural network satisfies fairness if it is the case; (2) otherwise, provide insights on why fairness is not satisfied and how fairness can be potentially achieved; (3) provide a way of improving the fairness of the neural network. At a high-level, our approach is designed as follows. Given a neural network (i.e., either a feed-forward or recurrent neural network), we systematically sample behaviors of the neural network (e.g., input/output pairs), based on which we learn a Markov Chain model that approximates the neural network. Our algorithm guarantees that probabilistic analysis based on the learned Markov Chain model (such as probabilistic reachability analysis) is probably approximately correct (hereafter PAC-correct) with respect to any computational tree logic (CTL~\cite{DBLP:journals/acta/Ben-AriPM83}) formulae. With the guarantee, we are thus able to verify fairness property of the neural network. There are two outcomes. One is that the neural network is proved to be fair, in which case the Markov Chain is presented as an evidence. Otherwise, sensitivity analysis based on the Markov Chain is carried out automatically. Such analysis helps us to understand why fairness is violated and provide hints on how the neural network could be improved to achieve fairness. Lastly, our approach optimizes the parameters of the `responsible' neurons in the neural network and improve its fairness.

We have implemented our approach as a part of the SOCRATES framework~\cite{socrates}. 
We apply our approach to multiple neural network models (including feed-forward and recurrent neural networks) trained on benchmark datasets which are the subject of previous studies on fairness testing. The experiment results show that our approach successfully verifies or falsifies all the models. It also confirms that fairness is a real concern and one of the networks (on the German Credit dataset) fails the fairness property badly. Through sensitivity analysis, our approach locates neurons which have the most contribution to the violation of fairness. Further experiments show that by optimizing the neural parameters (i.e., weights) based on the sensitivity analysis result, we can improve the model's fairness significantly whilst keeping a high model accuracy.

The remaining of the paper is organized as follows. In Section~\ref{sec:preli}, we review relevant background and define our problem. In Section~\ref{sec:approach}, we present each step of our approach in detail. In Section~\ref{sec:imple}, we evaluate our approach through multiple experiments. We review related work in Section~\ref{sec:related} and conclude in Section~\ref{sec:conclusion}. 

\section{Preliminary} \label{sec:preli}
In this section, we review relevant background and define our problem. \\

\noindent \emph{Fairness} For classification problems, a neural network $N$ learns to predict a target variable $O$ based on a set of input features $X$. We write $Y$ as the prediction of the classifier. We further write $F \subseteq X$ as a set of features encoding some protected characteristics such as gender, age and race. Fairness constrains how $N$ makes predictions. 
In the literature, there are multiple formal definitions of fairness~\cite{demographic,verification,fairness,counterfactual}. In this work, we focus on independence-based fairness, which is defined as follows.

\begin{definition}[Independence-based Fairness (strict)] \label{def:strict}
A neural network $N$ satisfies independence-based fairness (strict) if the protected feature $F$ is statistically independent to the prediction $Y$. We write $L$ as the prediction set and we have $\forall l \in L, \; \forall f_i, f_j \in F \: such~that~\: i \neq j$,
\begin{equation}
\begin{aligned}
P(Y = l \: | \: F = f_i)=P(Y = l \: | \: F = f_j)
\end{aligned}
\end{equation}
\end{definition}
The definition states that, 
$N$'s prediction is independent of the protected feature $F$. 
This definition is rather strict and thus unlikely to hold in practice. The following relaxes the above definition by introducing a positive tolerance $\xi$.
\begin{definition}[Independence-based Fairness]
\label{def:fairness}
Let $N$ be a neural network and $\xi$ be a positive real-value constant. $N$ satisfies independence-based fairness, with respect to $\xi$, if and only if, $\forall l \in L \: \forall f_i, f_j \in F~such~that~i \neq j$,
\begin{equation}
\begin{aligned}
| \: P(Y=l \: | \: F=f_i)-P(Y=l \: | \: F=f_j) \: | \: \leq \xi
\end{aligned}
\end{equation}
\end{definition}
Intuitively, the above definition states that $N$ is fair as long as the probability difference is within the threshold $\xi$. In the following, we focus on Definition~\ref{def:fairness} as it is both more general and more practical compared to Definition~\ref{def:strict}.

\begin{example} \label{example}
Let us take the network trained on the Census Income dataset~\cite{census} as an example. The dataset consists of 32k training instances, each of which contains 13 features. The task is to predict whether an individual's income exceeds \$50K per year. 
An example instance $x$ with a prediction $y$ will be $x : \langle 3 \; 5 \; 3 \; 0 \; 2 \; 8 \; 3 \; 0 \; \mathbf{1} \; 2 \; 0 \; 40 \; 0 \rangle$, $y : \langle 0 \rangle$.
Note that all features are categorical (i.e., processed using binning). Among all features, gender, age and race are considered protected features. 
The model $N$ trained based on the dataset is in the form of a six-layer fully-connected feed-forward neural network. The following is a fairness property defined based on the protected feature gender. 
\begin{equation}
\begin{aligned}
| \: P(Y=1 \: | \: F=male)-P(Y=1 \: | \: F=female) \: | \: \leq 0.1 
\end{aligned}
\end{equation}
Intuitively, the difference in the probability of predicting 1, for males and females, should be no more than 10\%. 
\end{example}

\noindent \emph{Our Problem} We are now ready to define our problem.

\begin{definition}[The verification problem]
Let $N$ be a neural network. Let $\phi$ be an independence-based fairness property (with respect to protected feature $F$ and a threshold $\xi$). The fairness verification problem is to verify whether $N$ satisfies $\phi$ or not. 
\end{definition}
One way of solving the problem is through statistical model checking (such as hypothesis testing~\cite{smc}). 
Such an approach is however not ideal. While it is possible to conclude whether $N$ is fair or not (with certain level of statistical confidence), the result often provides no insight. In the latter case, we would often be interested in performing further analysis to answer questions such as whether certain feature or neuron at a hidden layer is particularly relevant to the fairness issue and how to improve the fairness. The above-mentioned approach offers little clue to such questions.

\section{Our Approach} \label{sec:approach}
In this section, we present details of our approach. Our approach is shown in Algorithm~\ref{alg:analysis}.
The first step is to learn a Markov Chain $D$ which guarantees that probabilistic analysis such as probabilistic reachability analysis based on $D$ is PAC-correct with respect to $N$. The second step is to apply probabilistic model checking~\cite{pmc} to verify $D$ against the fairness property $\phi$. In the third step, if the property $\phi$ is not verified, sensitivity analysis is performed on $D$ which provides us information on how to improve $N$ in terms of fairness. That is, we improve the fairness of the model by optimizing the neuron weights based on the sensitivity analysis results. 

Note that our approach relies on building an approximation of the neural network in the form of Markov Chains. There are three reasons why constructing such an abstraction is beneficial. First, it allows us to reason about unbounded behaviors (in the case of a cyclic Markov Chains, which can be constructed from recurrent neural networks as we show below) which are known to be beyond the capability of statistical model checking~\cite{smc}. Second, the Markov Chain model allows us to perform analysis such as sensitivity analysis (e.g., to identify neurons responsible for violating fairness) as well as predict the effect of changing certain probability distribution (e.g., whether fairness will be improved), which are challenging for statistical methods. Lastly, in the case that the fairness is verified, the Markov Chain serves as a human-interpretable argument on why fairness is satisfied.  

In the following, we introduce each step in detail. We fix a neural network $N$ and a fairness property $\phi$ of the form $| \: P(Y=l \: | \: F=f_i)-P(Y=l \: | \: F=f_j) \: | \: \leq \xi$.  We use the neural network trained on the Census Income dataset (refer to Example~\ref{example}) as a running example.

\begin{minipage}[t]{0.55\linewidth}
\begin{algorithm}[H]
\DontPrintSemicolon
  \footnotesize
    Fix the set of states $S$;\\
    Learn DTMC $D$ by $learn(N, S, \frac{\mu\epsilon}{2}, 1 - \sqrt{1-\mu\delta})$; \\ 
    Estimate $P(Y=l \: | \: F=f_i)$ $\forall f_i \in F$; \\
    Verify $\phi$ against $\xi$;
    \If{$\phi$ is verified}
    {
        \Return{``Verified'' and $D$;}
    }
    \Else
    {
        Conduct sensitivity analysis on $D$; \\
        Perform automatic repair of $N$;\\
        \Return{$N'$;}
    }
    
\caption{$verify\_repair(N, \phi, \mu\epsilon, \mu\delta)$}
\label{alg:analysis}
\end{algorithm}
\end{minipage}
\hfill
\begin{minipage}[t]{0.4\linewidth}
\begin{algorithm}[H]
\DontPrintSemicolon
\footnotesize
    $W:=0$;\\
    $A_W:=0$;\\
    
    \Do{$\exists p \in S, n_p < H(n)$}
    {
        generate new sample trace $\omega$\\
        $W := W + \omega$;\\
        update $A_W(p,q)$
        for all $p \in S$ and $q \in S$;\\
        update $H(n)$;\\
    }
    \KwOutput{$A_W$}
\caption{learn($N, S, \epsilon, \delta$)}
\label{alg:learn}
\end{algorithm}
\end{minipage}

\subsection{Step 1: Learning a Markov Chain} \label{sec:step1}
In this step, we construct a Discrete-Time Markov Chain (DTMC) which approximates $N$ (i.e., line 2 of Algorithm~\ref{alg:analysis}). 
DTMCs are widely used to model the behavior of stochastic systems~\cite{DBLP:conf/cav/BazilleGJS20}, and they are often considered reasonably human-interpretable. Example DTMCs are shown in Figure~\ref{fig:example}. The definition of DTMC is presented in Appendix~\ref{A:DTMC}. Algorithm~\ref{alg:learn} shows the details of this step. The overall idea is to construct a DTMC, based on which we can perform various analysis such as verifying fairness. To make sure the analysis result on the DTMC applies to the original $N$, it is important that the DTMC is constructed in such a way that it preserves properties such as probabilistic reachability analysis (which is necessary for verifying fairness as we show later). Algorithm~\ref{alg:learn} is thus base on the recent work published in~\cite{DBLP:conf/cav/BazilleGJS20}, which develops a sampling method for learning DTMC. 
To learn a DTMC which satisfies our requirements, we must answer three questions.

\emph{(1) What are the states $S$ in the DTMC?} The choice of $S$ has certain consequences in our approach. First, it constrains the kind of properties that we are allowed to analyze based on the DTMC. As we aim to verify fairness, the states must minimally include states representing different protected features, and states representing prediction outcomes. The reason is that, with these states, we can turn the problem of verifying fairness into probabilistic reachability analysis based on the DTMC, as we show in Section~\ref{sec:step2}. 
What additionally are the states to be included  depends on the level of details that we would like to have for subsequent analysis. For instance, we include states representing other features at the input layer, and states representing the status of hidden neurons.
Having these additional states allows us to analyze the correlation between the states and the prediction outcome. For instance, having states representing a particular feature (or the status of a neuron of a hidden layer) allows us to check how sensitive the prediction outcome is with respect to the feature (or the status of a neuron). Second, the choice of states may have an impact on the cost of learning the DTMC. In general, the more states there are, the more expensive it is to learn the DTMC. In Section~\ref{sec:imple}, we show empirically the impact of having different sizes of $S$. We remark that to represent continuous input features and hidden neural states using discrete states, we discretize their values (e.g., using binning or clustering methods such as \emph{Kmeans}~\cite{kmeans} based on a user-provided number of clusters).  

\emph{(2) How do we identify the transition probability matrix?} The answer is to repeatedly sample inputs (by sampling based on a prior probability distribution) and then monitor the trace of the inputs, i.e., the sequence of transitions triggered by the inputs. After sampling a sufficiently large number of inputs, the transition probability matrix then can be estimated based on the frequency of transitions between states in the traces. In general, the question of estimating the transition probability matrix of a DTMC is a well-studied topic and many approaches have been proposed, including frequency estimation,  Laplace smoothing~\cite{DBLP:conf/cav/BazilleGJS20} and Good-Turing estimation~\cite{goodt}. In this work, we adopt the following simple and effective estimation method.
Let $W$ be a set of traces which can be regarded as a bag of transitions. We write $n_p$ where $p \in S$ to denote the number transitions in $W$ originated from state $p$. We write $n_{pq}$ where $p \in S$ and $q \in S$ to be the number of transitions observed from state $p$ to $q$ in $W$. Let $m$ be the total number of states in $S$. The transition probability matrix $A_W$ (estimated based on $W$) is: $    A_W(p,q)=\left\{
                \begin{array}{ll}
                  \frac{n_{pq}}{n_p} & \mbox{if $n_q \neq 0$} \\
                  \frac{1}{m} & \mbox{otherwise}
                \end{array}
              \right.$.
 Intuitively, the probability of transition from state $p$ to $q$ is estimated as the number of transitions from $p$ to $q$ divided by the total number of transitions taken from state $p$ observed in $W$. Note that if a state $p$ has not been visited, $A_W(p,q)$ is estimated by $\frac{1}{m}$; otherwise, $A_W(p,q)$ is estimated by $\frac{n_{pq}}{n_p}$.
 
 \emph{(3) How do we know that the estimated transition probability matrix is accurate enough for the purpose of verifying fairness?} Formally, let $A_W$ be the transition probability matrix estimated as above; and let $A$ be the actual transition probability matrix. We would like the following to be satisfied.  
\begin{equation}
P(\emph{Div}(A, A_W) > \epsilon) \leq \delta
\end{equation}
where $\epsilon > 0$ and $\delta > 0$ are constants representing \emph{accuracy} and \emph{confidence}; $\emph{Div}(A, A_W)$ represents the divergence between $A$ and $A_W$; and $P$ is the probability. Intuitively, the learned DTMC must be estimated such that the probability of the divergence between $A_W$ and $A$ greater than $\epsilon$ is no larger than the confidence level $\delta$. 
In this work, we define the divergence based on the individual transition probability, i.e.,
\begin{equation}
    P(\exists p \in S, \sum_{q \in S}{\big|A(p,q) - A_W(p,q)\big|} > \epsilon) \leq \delta
\end{equation}
Intuitively, we would like to sample traces until the observed transition probabilities $A_W(p,q)= \frac{n_{pq}}{n_p}$ are close to the real transition probability $A(p,q)$ to a certain level for all $p,q \in S$. Theorem 1 in the recently published work~\cite{DBLP:conf/cav/BazilleGJS20} shows that if we sample enough samples such that for each $p \in S$, $n_p$ satisfies
\begin{equation}
\label{h(n)}
    n_{p} \geq \frac{2}{\epsilon^2}log(\frac{2}{\delta'}) \Big[ \frac{1}{4} - \Big( max_q\Big|\frac{1}{2} - \frac{n_{pq}}{n_p}\Big| - \frac{2}{3}\epsilon \Big)^2 \Big]
\end{equation}
where $\delta' = \frac{\delta}{m}$, we can guarantee the learned DTMC is sound with respect to $N$ in terms of probabilistic reachability analysis. Formally, let $H(n) = \frac{2}{\epsilon^2}log(\frac{2}{\delta'}) [ \frac{1}{4} - (max_q|\frac{1}{2} - \frac{n_{pq}}{n_p}| - \frac{2}{3}\epsilon)^2]$,
\begin{theorem} \label{main}
Let $(S, I, A_W)$ be a DTMC where $A_W$ is the transition probability matrix learned using frequency estimation based on $n$ traces W. For $\;0 < \epsilon < 1 \; and \; 0 < \delta < 1$, if for all $p \in S, \; n_p \geq H(n)$, we have for any CTL property $\psi$,
\begin{equation}
P(\big| \gamma(A, \psi) - \gamma(A_W, \psi)\big | > \epsilon) \leq \delta
\end{equation}
where $\gamma(A_W, \psi)$ is the probability of $A_W$ satisfying $\psi$. 
\end{theorem}

Appendix~\ref{A:proof} provides the proof. Intuitively, the theorem provides a bound on the number of traces that we must sample in order to guarantee that the learned DTMC is PAC-correct with respect to any CTL property, which provides a way of verifying fairness as we show in Section~\ref{sec:step2}. 

We now go through Algorithm~\ref{alg:learn} in detail. The loop from line 3 to 8 keeps sampling inputs and obtains traces. Note that we use the uniform sampling by default and would sample according to the actual distribution if it is provided. 
Next, we update  $A_W$ as explained above at line 6. Then we check if more samples are needed by monitoring if a sufficient number of traces has been sampled according to Theorem~\ref{main}. If it is the case, we output the DTMC as the result. Otherwise, we repeat the steps to generate new samples and update the model. 
\begin{example}
In our running example, for simplicity assume that we select \emph{gender} (as the protected feature) and the prediction outcome to be included in $S$ and the number of clusters is set to 2 for both layers. Naturally, the two clusters identified for the protected feature are male and female (written as \emph{`M'} and \emph{`F'}) and the two clusters determined for the outcome are \emph{$`\leq 50K'$} and \emph{$`> 50K'$}. A sample trace is $w = \langle Start, `M', `> 50K' \rangle $, where $Start$ is a dummy state where all traces start. 
Assume that we set accuracy $\epsilon = 0.005$ and confidence level $\delta = 0.05$. Applying Algorithm~\ref{alg:learn}, 2.85K traces are generated to learn the transition matrix $A_W$.  
The learned DTMC $D$ is shown in Figure~\ref{fig:gen}. 
\end{example}

    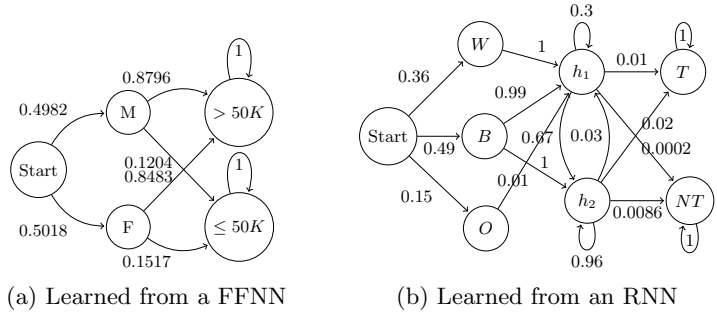
\begin{figure}[t]
    \centering
    \subfloat[\centering Learned from a FFNN]{
        \resizebox{0.3\textwidth}{!}{
        \begin{tikzpicture}
            \node[state]             (s0) {Start};
            \node[state, above right=of s0,  yshift=-0.6cm] (input1) {M};
            \node[state, below right=of s0,  yshift=0.6cm] (input2) {F};
            \node[state, right=of input1] (output1) {$> 50K$};
            \node[state, right=of input2] (output2) {$\leq 50K$};
    
            \draw[every loop]
                (s0) edge[bend left, auto=left]  node {$0.4982$} (input1)
                (s0) edge[bend right, auto=right] node {$0.5018$} (input2)
                (input1) edge[bend left, auto=left]  node {$0.8796$} (output1)
                (input1) edge[auto=right]  node {$0.8483$} (output2)            
                (input2) edge[auto=left] node {$0.1204$} (output1)
                (input2) edge[bend right, auto=right] node {$0
                .1517$} (output2)            
                (output1) edge[loop above][auto=right]             node {1} (output1)
                (output2) edge[loop above][auto=right]           node {1} (output2);
        \end{tikzpicture}
        }
       \label{fig:gen}
    }%
    \qquad
    \subfloat[\centering Learned from an RNN]{
        \resizebox{0.4\textwidth}{!}{
        \begin{tikzpicture}
            \node[state]             (s0) {Start};
            \node[state, above right=of s0,  yshift=0cm] (input1) {$W$};
            \node[state, right=of s0,  xshift=-0.2cm] (input2) {$B$};
            \node[state, below right=of s0,  xshift=0.1cm] (input3) {$O$};
            \node[state, right=of input1, yshift=-0.5cm] (hidden1) {$h_1$};
            \node[state, right=of input3, yshift=0.5cm] (hidden2) {$h_2$};
            \node[state, right=of hidden1] (output1) {$T$};
            \node[state, right=of hidden2] (output2) {$NT$};
    
            \draw[every loop]
                (s0) edge[auto=left]  node {0.36} (input1)
                (s0) edge[auto=right] node {0.49} (input2)
                (s0) edge[auto=right] node {0.15} (input3)
                (input1) edge[auto=left]  node {1} (hidden1)
                (input2) edge[auto=left] node {0.99} (hidden1)
                (input2) edge[auto=right] node {0.01} (hidden2)  
                (input3) edge[auto=right] node {1} (hidden1)
                (hidden1) edge[bend right, auto=right] node {0.67} (hidden2)  
                (hidden2) edge[bend right, auto=left] node {0.03} (hidden1)
                (hidden1) edge[loop above][auto=above]           node {0.3} (hidden1)
                (hidden2) edge[loop below][auto=below]           node {0.96} (hidden2)
                (hidden1) edge[auto=left]  node {0.01} (output1)
                (hidden1) edge[auto=left]  node {0.02} (output2)            
                (hidden2) edge[auto=right] node {0.0002} (output1)
                (hidden2) edge[auto=right] node {0.0086} (output2)  
                (output1) edge[loop above][auto=right]           node {1} (output1)
                (output2) edge[loop below][auto=right]           node {1} (output2);
        \end{tikzpicture}
        }
         \label{fig:rnn}
    }%
    \caption{Sample learned DTMCs}%
    \label{fig:example}%
\end{figure}

\subsection{Step 2: Probabilistic Model Checking} \label{sec:step2}
In this step, we verify $N$ against the fairness property $\phi$ based on the learned $D$. Note that $D$ is PAC-correct only with respect to CTL properties.
Thus it is infeasible to directly verify $\phi$ (which is not a CTL property). Our remedy is to compute $P(Y=l \: | \: F=f_i)$ and $P(Y=l \: | \: F=f_j)$ separately and then verify $\phi$ based on the results. Because we demand there is always a state in $S$ representing $F=f_i$ and a state representing $Y=l$, the problem of computing $P(Y=l \: | \: F=f_i)$ can be reduced to a probabilistic reachability checking problem $\psi$, i.e., the probability of reaching the state representing $Y=l$ from the state representing $F=f_i$. This can be solved using probabilistic model checking techniques.     
Probabilistic model checking~\cite{pmc} of DTMC is a formal verification method for analyzing DTMC against formally-specified quantitative properties (e.g., PCTL). 
Probabilistic model checking algorithms are often based on a combination of graph-based algorithms and numerical techniques. For straightforward properties such as computing the probability that a $\textbf{U}$ (Until), $\textbf{F}$ (Finally) or $\textbf{G} (Globally)$ path formula is satisfied, the problem can be reduced to solving a system of linear equations~\cite{pmc}. We refer to the readers to~\cite{pmc} for a complete and systematic formulate of the algorithm for probabilistic model checking.

\begin{example}

Figure~\ref{fig:rnn} shows a DTMC learned from a recurrent neural network trained on Jigsaw Comments dataset (refer to details on the dataset and network in Section \ref{sec:dataset}). The protected features is \emph{race}. For illustration purpose, let us consider three different values for \emph{race}, i.e., White (\emph{W}), Black (\emph{B}) and Others (\emph{O}). For the hidden layer cells, we consider LSTM cell 1 only and cluster its values into two groups, represented as two states $h_1$ and $h_2$. The output has two categories, i.e., Toxic (\emph{T}) and Non-Toxic (\emph{NT}). The transition probabilities are shown in the figure. Note that the DTMC is cyclic due to the recurrent hidden LSTM cells in the network. 
We obtained $P(Y=\emph{`T'} \:| \:F=\emph{`W'})$ by probabilistic model checking as discussed above. 
The resultant probability is $0.0263$. Similarly, $P(Y=\emph{`T'} \:| \:F=\emph{`B'})$ and $P(Y=\emph{`T'} \:| \:F=\emph{`O'})$ are $0.0362$ and $0.0112$ respectively.
\end{example}

Next we verify the fairness property $\phi$ based on the result of probabilistic model checking. First, the following is immediate based on Theorem~\ref{main}.

\begin{proposition}
Let $D = (S, I, A_W)$ be a DTMC learned using Algorithm~\ref{alg:learn}. Let $P(Y=l \: | \: F=f_i)$ be the probability computed based on probabilistic model checking $D$ and $P_t(Y=l \: | \: F=f_i)$ is the actual probability in $N$. We have
\begin{center}
$P\big(\big |P(Y=l \: | \: F=f_i) - P_t(Y=l \: | \: F=f_i)\big| > \epsilon\big) \leq \delta$ 
\end{center}
\end{proposition}

\begin{theorem}
Let $X$ be an estimation of a probability $X_t$ such that $P(|X - X_t| > \epsilon) \leq \delta$. Let $Z$ be an estimation of a probability $Z_t$ such that $P(|Z - Z_t| > \epsilon) \leq \delta$. We have 
$P(|(X - Z)-(X_t - Z_t)| > 2\epsilon) \leq 2\delta - \delta^2$. 
\label{the:probdiff}
\end{theorem}
Appendix~\ref{A:proofprobdiff} provides the proof. Hence, given an expected accuracy $\mu\epsilon$ and a confidence level $\mu\delta$ on fairness property $\phi$
, we can derive $\epsilon$ and $\delta$ to be used in Algorithm 2 as: $\epsilon = \frac{\mu\epsilon}{2}$ and $\delta = 1-\sqrt{1-\mu\delta}$.
We compute the probability of $P(Y=l \: | \: F=f_i)$ and $P(Y=l \: | \: F=f_j)$ based on the learned $D$ (i.e., line 3 of Algorithm~\ref{alg:analysis}). Next, we compare $| P(Y=l \: | \: F=f_i) - P(Y=l \: | \: F=f_j) |$ with $\xi$. If the difference is no larger than $\xi$, fairness is verified. The following establishes the correctness of Algorithm~\ref{alg:analysis}.

\begin{theorem}\label{the:pac}
Algorithm~\ref{alg:analysis} is PAC-correct with accuracy $\mu\epsilon$ and confidence $\mu\delta$, if Algorithm 2 is used to learn the DTMC $D$.
\end{theorem}
Appendix~\ref{A:proof3} provides the proof. 

The overall time complexity of model learning and probabilistic model checking is linear in the number of traces sampled, i.e., $\textbf{O}(n)$ where $n$ is the total number of traces sampled. Here $n$ is determined by $H(n)$ as well as the probability distribution of the states. Contribution of $H(n)$ can be determined as $\textbf{O}(\frac{\log m}{\mu\epsilon^2\log \mu\delta})$ based on Equation~\ref{h(n)},
where $m$ is the total number of states.
In the first case, for a model with only input features and output predictions as states, the probability of reaching each input states are statistically equal if we apply uniform sampling to generate IID input vectors. In this scenario the overall time complexity is $\textbf{O}(\frac{m\log m}{\mu\epsilon^2\log \mu\delta})$. In the second case, for a model with states representing the status of hidden layer neurons, we need to consider the probability for each hidden neuron states when the sampled inputs are fed into the network $N$. In the best case, the probabilities are equal, we denote $m'$ as the maximum number of states in one layer among all layers included, the complexity is then $\textbf{O}(\frac{m'\log m}{\mu\epsilon^2\log \mu\delta})$. In the worst case, certain neuron is never activated (or certain predefined state is never reached) no matter what the input is. Since the probability distribution among the hidden states are highly network-dependent, we are not able to estimate the average performance. 

\begin{example}
In our running example, with the learned $A_W$ of $D$ as shown in Figure~\ref{fig:gen}, the probabilities as $\: P(Y=1 \: | F = \emph{`F'}) = 0.8483$ and $\: P(Y=1 \: | F = \emph{`M'}) = 0.8796$.
Hence, $| \: P(Y=1 \: | F = \emph{`F'})-P(Y=1 \: | F = \emph{`M'}) \: |  = 0.0313$. 
Next, we compare the probability difference against the user-provided fairness criteria $\xi$. If $\xi = 0.1$, $N$ satisfies fairness property. If $\xi = 0.02$, $N$ fails fairness. Note that such a strict criteria is not practical and is used for illustration purpose only. 
\end{example}

\subsection{Step 3: Sensitivity Analysis} \label{sec:step3}
In the case that the verification result shows $\phi$ is satisfied, our approach outputs $D$ and terminates successfully. We remark that in such a case $D$ can be regarded as the evidence for the verification result as well as a human-interpretable explanation on why fairness is satisfied. In the case that $\phi$ is not satisfied, a natural question is: how do we improve the neural network for fairness? Existing approaches have proposed methods for improving fairness such as by training without the protected features~\cite{FairerML} (i.e., a form of pre-processing) or training with fairness as an objective~\cite{ftrain1} (i.e., a form of in-processing). In the following, we show that a post-processing method can be supported based on the learned DTMC. That is, we can identify the neurons which are responsible for violating the fairness based on the learned DTMC and ``mutate'' the neural network slightly, e.g., adjusting its weights, to achieve fairness. 

We start with a sensitivity analysis to understand the impact of each probabilistic distribution (e.g., of the non-protected features or hidden neurons) on the prediction outcome. Let $F$ be the set of discrete states representing different protected feature values. Let $I$ represent a non-protected feature or an internal neuron. We denote $I_i$ as a particular state in the DTMC which represents certain group of values of the feature or neuron. Let $l$ represent the prediction result that we are interested in. The sensitivity of $I$ (with respect to the outcome $l$) is defined as follows. 
\begin{equation*} \label{sensitivity}
sensitivity(I) = \sum_{i}reach(S_0, I_i) * reach(I_i, l) * \max_{\{f,g\} \subseteq F} \big (reach(f, I_i) - reach(g, I_i) \big )
\end{equation*}
where $reach(s, s')$ for any state $s$ and $s'$ represents the probability of reaching $s'$ from $s$.
Intuitively, the sensitivity of $I$ is the summation of the `sensitivity' of every state $I_i$, which is calculated as $\max_{f, g} \big (reach(f, I_i) - reach(g, I_i) \big )$, i.e., the maximum probability difference of reaching $I_i$ from all possible protected feature states. The result is then multiplied with the probability of reaching $I_i$ from start state $S_0$ and the probability of reaching $l$ from $I_i$. Our approach analyzes all non-protected features and hidden neurons and identify the most sensitive features or neurons for improving fairness in step 4.

\begin{example}
In our running example, based on the learned DTMC $D$ shown in Figure~\ref{fig:gen}, we perform sensitivity analysis as discussed above.
We observe that feature 9 (i.e., representing `capital gain') is the most sensitive, i.e., it has the most contribution to the model unfairness. More importantly, it can be observed that the sensitivities of the neurons vary significantly, which is a good news as it suggests that for this model, optimizing the weights of a few neurons may be sufficient for achieving fairness. Figure~\ref{fig:seneg} in Appendix~\ref{A.eg} shows the sensitively analysis scatter plot.
\end{example}

\subsection{Step 4: Improving Fairness} \label{sec:step4}
In this step, we demonstrate one way of improving neural network fairness based on our analysis result, i.e., by adjusting weight parameters of the neurons identified in step 3. The idea is to search for a small adjustment through optimization techniques such that the fairness property is satisfied. In particular, we adopt the Particle Swarm Optimization (PSO) algorithm~\cite{pso1995}, which simulates intelligent collective behavior of animals such as flocks of birds and schools of fish. In PSO multiple particles are placed in the search space and the optimization target is to find the best location, where the fitness function is used to determine the best location. We omit the details of PSO here due to space limitation and present it in Appendix~\ref{A:PSO}.

In our approach, the weights of the most sensitive neurons are the subject for optimization and thus are represented by the location of the particles in the PSO. The initial location of each particle is set to the original weights and the initial velocity is set to zero.
The fitness function is defined as follows.
\begin{equation}
    fitness = Prob_{diff} + \alpha (1 - accuracy)
\end{equation}
where $Prob_{diff}$ represents the maximum probability difference of getting a desired outcome among all different values of the sensitive feature; $accuracy$ is the accuracy of repaired network on the training dataset and constant parameter $\alpha \in (0,1)$ determines the importance of the accuracy (relative to the fairness). Intuitively, the objective is to satisfy fairness and not to sacrifice accuracy too much.  We set the bounds of weight adjustment to $(0,2)$, i.e., 0 to 2 times of the original weight. The maximum number of iteration is set to 100. To further reduce the searching time, we stop the search as soon as the fairness property is satisfied or we fail to find a better location in the last 10 consecutive iterations.

\begin{example}
In our running example, we optimize the weight of ten most sensitive neurons
using PSO for better fairness. The search stops at the $13^{rd}$ iteration as no better location is found in the last 10 consecutive iterations. The resultant probability difference among the protected features dropped from $0.0313$ to $0.007$, whereas the model accuracy dropped from $0.8818$ to $0.8606$.
\end{example}

\section{Implementation and Evaluation} \label{sec:imple}
Our approach has been implemented on top of SOCRATES~\cite{socrates}, which is a framework for experimenting neutral network analysis techniques. 
We conducted our experiments on a machine with 1 Dual-Core Intel Core i5 2.9GHz CPU and 8GB system memory.

\subsection{Experiment Setup} \label{sec:dataset}
In the following, we evaluate our method in order to answer multiple research questions (RQs) based on multiple neural networks trained on 4 datasets adopted from existing research~\cite{aequitas2018,SG,adf2020,counterfactual}, i.e., in addition to the \emph{Census Income~\cite{census}} dataset as introduced in Example~\ref{example}, we have the following three datasets.
First is the \emph{German Credit}~\cite{credit} dataset consisting of 1k instances containing 20 features and is used to assess an individual's credit. Age and gender are the two protected features. The labels are whether an individual's credit is good or not. Second is the \emph{Bank Marketing~\cite{bank}} dataset consisting of 45k instances. There are 17 features, among which age is the protected feature. The labels are whether the client will subscribe a term deposit. Third is Jigsaw Comment~\cite{jigsaw} dataset. It consists of 313k text comments with average length of 80 words classified into toxic and non-toxic. The protected features analysed are race and religion.

Following existing approaches~\cite{aequitas2018,SG,adf2020,counterfactual}, we train three 6-layer feed-forward neural networks (FFNN) on the first three dataset (with accuracy 0.88, 1 and 0.92 respectively) and train one recurrent neural network, i.e., 8-cell Long Short Term Memory (LSTM), for the last dataset (with accuracy 0.92) and analyze their fairness against the corresponding protected attributes. 
For the LSTM model, we adopt the state-of-the-art embedding tool GloVe~\cite{glove}. We use the 50-dimension word vectors pre-trained on Wikipedia 2014 and Gigaword 5 dataset. 

Recall that we need to sample inputs to learn a DTMC. In the case of first three datasets, inputs are sampled by generating randomly values within the range of each feature (in IID manner assuming a uniform distribution). In the case of the Jigsaw dataset, we cannot randomly generate and replace words as the resultant sentence is likely invalid. Inspired by the work in~\cite{textbugger,synonym_1,synonym_2}, our approach is to replace a randomly selected word with a randomly selected synonym (generated by Gensim~\cite{gensim}). 

\begin{table}[t]
\centering
\caption{Fairness Verification Results}
\begin{tabular}{c | c | c | c | c | c | c}
\toprule
Dataset&Feature&\#States&\#Traces&Max Prob. Diff.& Result& Time\\
\midrule
Census&Race&8&12500&0.0588&PASS&4.13s\\
Census&Age&12&23500&0.0498&PASS&6.31s\\
Census&Gender&5&2850&0.0313&PASS&0.98s\\
Credit&Age&11&22750&0.1683&Fail&6.72s\\
Credit&Gender&5&2850&0.0274&PASS&1.01s\\
Bank&Age&12&27200&0.0156&PASS&6.33s\\
Jigsaw&Religion&10&35250&0.0756&PASS&29.6m\\
Jigsaw&Race&7&30550&0.0007&PASS&27.3m\\
\bottomrule
\end{tabular}
\label{tab:rq1}
\end{table}

\subsection{Research Questions and Answers}
\noindent \emph{RQ1: Is our approach able to verify fairness?} 
We systematically apply our method to the above-mentioned neural networks with respect to each protected feature. Our experiments are configured with accuracy $\mu\epsilon = 0.01$, confidence level $\mu\delta = 0.1$ (i.e., $\epsilon = 0.005$, $\delta = 0.05$) and fairness criteria $\xi = 10\%$ (which is a commonly adopted threshold~\cite{fairsquare}). Furthermore, in this experiment, the states in the DTMC $S$ are set to only include those representing the protected feature and different predictions. Table~\ref{tab:rq1} summarizes the results.
We successfully verify or falsify all models. Out of eight cases, the model trained on the German Credit dataset fails fairness with respect to the feature \emph{age} (i.e., the maximum probability difference among different age groups is 0.1683 which is greater than $\xi = 10\%$). 
Furthermore, the model trained on the Jigsaw dataset shows some fairness concern with respect to the feature \emph{religion} (although the probablity different is still within the threshold). This result shows that fairness violation could be a real concern.

\noindent \emph{RQ2: How efficient is our approach?} 
We answer the question using two measurements. The first measurement is the execution time. The results are shown in the last column in Table~\ref{tab:rq1}. For the six cases of FFNN, the average time taken to verify a network is around 4.25s, with a maximum of 6.72s for the model trained on German Credit on feature \emph{age} and a minimum of 0.98 seconds for the model trained on the Census Income dataset on feature \emph{gender}. For the two cases of LSTM networks, the average time taken is 30 minutes. Compared with FFNN, verifying an LSTM requires much more time. This is due to three reasons. Firstly, as mentioned in Section~\ref{sec:dataset}, sampling texts requires searching for synonyms. This is non-trivial due to the large size of the dictionary. 
Secondly, during sampling, we randomly select instances from the training set and apply perturbation to them in order to generate new samples. 
However, most of the instances in the Jigsaw training set does not contain the sensitive word. This leads to an increased number of traces needed to learn a DTMC. Thirdly, the LSTM model takes much more time to make a prediction than that by FFNN in general.
It is also observed that for all the cases, the execution time is proportional to the number of traces used in DTMC model learning (as discussed in our time complexity analysis). 
\begin{table}
	\begin{minipage}{0.5\linewidth}
    \centering

        \begin{tabular}{|p{1.2cm}|p{2.4cm}|p{2.4cm}|}
        \hline 
        Dataset & Max Probability Difference & Accuracy \\
        \hline
        German &0.1683 $\mapsto$ 0.1125 & 1.0 $\mapsto$ 0.9450\\
        Census & 0.0588 $\mapsto$ 0.0225 & 0.8818 $\mapsto$ 0.8645\\
        Jigsaw & 0.0756 $\mapsto$ 0.0590 & 0.9166 $\mapsto$ 0.9100\\
        \hline
        \end{tabular}
        \caption{Fairness Improvement}
        \label{tab:imp}
	\end{minipage}\hfill
	\begin{minipage}{0.45\linewidth}
       \begin{tikzpicture}
                \pgfplotsset{%
                    width=1\textwidth,
                    height=0.6\textwidth
                }
                \begin{axis}[
                    xlabel={\#States},
                    ylabel={Execution time (s)},
                    xmin=0, xmax=510,
                    ymin=0, ymax=1300,
                    xtick = {100, 200, 500},
                    ytick={100, 500, 1000, 1300},
                    legend pos=north west,
                    ymajorgrids=false,
                    grid style=dashed,
                ]
                
                \addplot[
                    mark=*,
                    mark size=0.8pt,
                    color = blue,
                    ]
                    coordinates {
                    (33,33.109419)(38,41.503351)(43,49.349741)(73,92.033411)(83,112.375049)(203,384.018105666667)(503,1278)
                    };
                    \legend{Analysis}
                
                \addplot[
                    dotted,
                    color = red,
                    ]
                    coordinates {
                    (33,50.1109600159703)(38,60.0317766714388)(43,70.2391435899222)(73,136.022568788793)(83,159.283481667214)(203,468.421695696382)(503,1358.88869648313)
                    };
                    \addlegendentry{$n\log n$}
                    
                \end{axis}
                \end{tikzpicture}
                
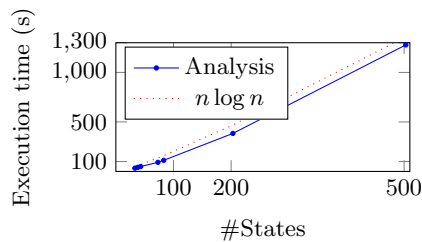
\captionof{figure}{Execution Times vs Number of States }
                \label{fig:time} 
	\end{minipage}
\end{table}
The other measurement is the number of traces that we are required to sample using Algorithm~\ref{alg:learn}. For each model and protected feature, the number of traces generated in Algorithm~\ref{alg:learn} depends on number of categorical values defined for this protected feature and the number of predictions. That is, more categories and predictions result in more states in the learn a DTMC model, which subsequently lead to more traces required. Furthermore, the number of traces required also depends on the probabilistic distribution from each state in the model. As described in Algorithm 2, the minimum number of traces transiting from each state must be greater than $H(n)$. This is evidenced by results shown in Table~\ref{tab:rq1}, where the number of traces vary significantly between models or protected features, ranging from 2K to 35K. Although the number of traces is expected to increase for more complicated models, we believe that this is not a limiting factor since the sampling of the traces can be easily paralleled.  

We further conduct an experiment to monitor the execution time required for the same neural network model with a different numbers of states in the learned DTMC. We keep other parameters (i.e., $\mu\epsilon$, $\mu\delta$ and $\phi$) the same. Note that hidden neurons are not selected as states to reduce the impact of the state distribution. We show one representative result (based on the mode trained on the Census Income dataset with attribute \emph{race} as the protected feature) in Figure~\ref{fig:time}. As we can see the total execution time is bounded by $n\log n$ which tally with our time complexity analysis in Section~\ref{sec:approach}. 


\noindent \emph{RQ3: Is our approach able to improve fairness and is the sensitivity analysis useful?} The question asks whether the sensitivity analysis results based on the learned DTMC can be used to improve fairness. 
To answer this question, we systematically perform sensitivity analysis (on both the input features and the hidden neurons) and optimize the weights of the neurons which are sensitive to fairness. 
We focus on three cases, i.e., the FFNN model trained on the German Credit model w.r.t \emph{age} and on the Census Income model w.r.t \emph{race} and the LSTM model trained on the Jigsaw comments w.r.t \emph{religion}, as the maximum probability difference for these three cases (as shown in Table~\ref{tab:rq1}) is concerning (i.e., $> 5\%$). For the former two, we optimize the weights of the top-10 sensitive neurons (including the first layer neurons representing other features). For the LSTM model, we optimize top-3 sensitive  cells (due to the small number of cells). 
Table~\ref{tab:imp} shows the fairness improvement as well as the drop in accuracy. It can be observed that in all three cases we are able to improve the fairness whilst maintaining the accuracy at a high-level. Note that the parameter $\alpha$ is set to $0.1$ in these experiments and it can be used to achieve better fairness or accuracy depending the user requirement.

\noindent \emph{RQ4: How does our approach compare with existing alternatives?} The most relevant tools that we identify are FairSquare~\cite{fairsquare} and VeriFair~\cite{verification}. FairSquare and VeriFair use numerical integration to verify fairness properties of machine learning models including neural networks. FairSquare relies on constraint solving techniques and thus it is difficult to scale to large neural networks. VeriFair is based on adaptive concentration inequalities. We evaluate our approach against these two tools on all eight models. For FairSquare and VeriFair, we follow the setting of independent feature distribution and check for demographic parity~\cite{verification}. For both tools, we set $c = 0.15$ as suggested and keep other parameters as default. As both FairSquare and VeriFair are designed to compare two groups of sub-populations, for those protected features that have more than two categories, we perform binning to form two groups.  For the six FFNN models, we set timeout value to be $900s$ following the setting in VeriFair. As shown in Table~\ref{tab:4.1}, FairSquare is not able to scale for large neural network and for all FFNN models it fails to verify or falsify the model in time. Both VeriFair and our approach successfully verified all six FFNN models. But our approach completes the verification within 1s for all models while VeriFair takes $62$ times more execution time than our approach on average. For the RNN models trained on Jigsaw dataset, neither FairSquare nor VeriFair is able to analyze them. FairSquare supports only loop-free models and, hence, it cannot handle RNN models. Although VeriFair is able to handle RNN networks in general, it does not support text classifiers. Hence, compared with existing solutions, our approach is more efficient than FairSquare and VeriFair and can support RNN-based text classifiers. 

\begin{table}[t]
\centering
\caption{Comparison with FairSquare and VeriFair}
\begin{tabular}{|cc|cc|cc|cc|}
\hline 
\multirow{2}{*}{Dataset} & \multirow{2}{*}{Prot.Feat.}&\multicolumn{2}{c|}{FairSquare}& \multicolumn{2}{c|}{VeriFair}& \multicolumn{2}{c|}{Ours}\\
&&Result&Time&Result&Time&Result&Time\\
\hline 
Census&Race&-&T.O.&Pass&2.33s&Pass&0.93s\\
Census&Age&-&T.O.&Pass&37.14s&Pass&0.81s\\
Census&Gender&-&T.O.&Pass&2.19s&Pass&0.89s\\
Credit&Age&-&T.O.&Pass&39.29s&Pass&0.90s\\
Credit&Gender&-&T.O.&Pass&8.23s&Pass&0.82s\\
Bank&Age&-&T.O.&Pass&245.34s&Pass&0.97s\\
Jigsaw&Religion&-&-&-&-&Pass&29.6m\\
Jigsaw&Race&-&-&-&-&Pass&27.3m\\

\hline 
\end{tabular}
\label{tab:4.1}
\end{table}

\section{Related Work} \label{sec:related}

\emph{Neural network verification.} 
There have been multiple approaches proposed to verify the robustness of neural networks utilizing various techniques, i.e., abstraction ~\cite{abstract20,deeppoly,ai2}, SMT sovler~\cite{marabou20,Reluplex,invariant20}, MILP and LP~\cite{Ehlers17,milp}, symbolic execution~\cite{symbolic18} and many others~\cite{bound20,ensemble20,exactness19,popqorn}. Besides~\cite{fairsquare} and~\cite{verification} that we addressed in RQ4, ~\cite{counterfactual} and~\cite{metamorphic} studied fairness property of text classifiers. Unlike ours, they focus on text classifiers only and their performance on RNN is unknown.

\emph{Fairness testing and improvement.} There have been an impressive line of methods proposed recently on machine learning model fairness testing and improvement. THEMIS~\cite{THEMIS,THEMIS_tool}, AEQUITAS~\cite{aequitas2018}, Symbolic Generation (SG)~\cite{SG} and ADF~\cite{adf2020}, are proposed to generate discriminatory instances for fairness testing. There are also existing proposals on fairness training~\cite{embedding,counterfactual,ftrain1,ftrain2,ftrain3,ftrain4}. Our work instead focuses on post-processing where a trained model is repaired based on sensitivity analysis results to improve fairness.

\emph{Machine learning model repair.} There have been multiple approaches proposed to repair machine learning models based on various technologies, i.e.,~\cite{patch19} leverages SMT solving,~\cite{minimal20} is based on advances in verification methods,~\cite{undertainty17} is guided by input population and etc. Unlike these methods, our work focuses on fairness repair and supports FFNN and RNN by design.

\section{Conclusion} \label{sec:conclusion}
In this work, we proposed an approach to formally verify neural networks against fairness properties. Our work relies on an approach for leaning DTMC from given neural network with PAC-correct guarantee. Our approach further performs sensitivity analysis on the neural network if it fails the fairness property and provides useful information on why the network is unfair. This result is then used as a guideline to adjust network parameters and achieve fairness. 
Comparing with existing methods evaluating neural network fairness, our approach has significantly better performance in terms of efficiency and effectiveness.

\section{Acknowledgements} \label{sec:ack}
We thank anonymous reviewers for their constructive feedback. This research is partly supported by project MOET 32020-0004 funded by Ministry of Education, Singapore.

%
%
%
%

\begin{thebibliography}{10}
\providecommand{\url}[1]{\texttt{#1}}
\providecommand{\urlprefix}{URL }
\providecommand{\doi}[1]{https://doi.org/#1}

\bibitem{jigsaw}
 (2017),
  \url{https://www.kaggle.com/c/jigsaw-toxic-comment-classification-challenge}

\bibitem{trust_ai}
Draft ethics guidelines for trustworthy {AI}. Tech. rep., European Commission
  (2018)

\bibitem{ftrain2}
Agarwal, A., Beygelzimer, A., Dud{\'{\i}}k, M., Langford, J., Wallach, H.M.: A
  reductions approach to fair classification. In: Dy, J.G., Krause, A. (eds.)
  Proceedings of the 35th International Conference on Machine Learning, {ICML}
  2018, Stockholmsm{\"{a}}ssan, Stockholm, Sweden, July 10-15, 2018.
  Proceedings of Machine Learning Research, vol.~80, pp. 60--69. {PMLR} (2018),
  \url{http://proceedings.mlr.press/v80/agarwal18a.html}

\bibitem{SG}
Agarwal, A., Lohia, P., Nagar, S., Dey, K., Saha, D.: Automated test generation
  to detect individual discrimination in {AI} models. CoRR  (2018),
  \url{http://arxiv.org/abs/1809.03260}

\bibitem{undertainty17}
Albarghouthi, A., D'Antoni, L., Drews, S.: Repairing decision-making programs
  under uncertainty. In: Majumdar, R., Kuncak, V. (eds.) Computer Aided
  Verification - 29th International Conference, {CAV} 2017, Heidelberg,
  Germany, July 24-28, 2017, Proceedings, Part {I}. Lecture Notes in Computer
  Science, vol. 10426, pp. 181--200. Springer (2017).
  \doi{10.1007/978-3-319-63387-9\_9},
  \url{https://doi.org/10.1007/978-3-319-63387-9\_9}

\bibitem{fairsquare}
Albarghouthi, A., D'Antoni, L., Drews, S., Nori, A.V.: Fairsquare:
  probabilistic verification of program fairness. Proc. {ACM} Program. Lang.
  \textbf{1}({OOPSLA}),  80:1--80:30 (2017). \doi{10.1145/3133904},
  \url{https://doi.org/10.1145/3133904}

\bibitem{synonym_1}
Alzantot, M., Sharma, Y., Elgohary, A., Ho, B., Srivastava, M.B., Chang, K.:
  Generating natural language adversarial examples. In: Proceedings of the 2018
  Conference on Empirical Methods in Natural Language Processing ({EMNLP}
  2018), Brussels, Belgium. pp. 2890--2896 (2018),
  \url{https://doi.org/10.18653/v1/d18-1316}

\bibitem{THEMIS_tool}
Angell, R., Johnson, B., Brun, Y., Meliou, A.: Themis: automatically testing
  software for discrimination. In: Proceedings of the 2018 {ACM} Joint Meeting
  on European Software Engineering Conference and Symposium on the Foundations
  of Software Engineering ({ESEC/SIGSOFT} {FSE} 2018), Lake Buena Vista, FL,
  USA. pp. 871--875 (2018), \url{https://doi.org/10.1145/3236024.3264590}

\bibitem{verification}
Bastani, O., Zhang, X., Solar{-}Lezama, A.: Probabilistic verification of
  fairness properties via concentration. {PACMPL}  \textbf{3}({OOPSLA}),
  118:1--118:27 (2019), \url{https://doi.org/10.1145/3360544}

\bibitem{DBLP:conf/cav/BazilleGJS20}
Bazille, H., Genest, B., J{\'{e}}gourel, C., Sun, J.: Global {PAC} bounds for
  learning discrete time markov chains. In: Lahiri, S.K., Wang, C. (eds.)
  Computer Aided Verification - 32nd International Conference, {CAV} 2020, Los
  Angeles, CA, USA, July 21-24, 2020, Proceedings, Part {II}. Lecture Notes in
  Computer Science, vol. 12225, pp. 304--326. Springer (2020).
  \doi{10.1007/978-3-030-53291-8\_17},
  \url{https://doi.org/10.1007/978-3-030-53291-8\_17}

\bibitem{DBLP:journals/acta/Ben-AriPM83}
Ben{-}Ari, M., Pnueli, A., Manna, Z.: The temporal logic of branching time.
  Acta Informatica  \textbf{20},  207--226 (1983). \doi{10.1007/BF01257083},
  \url{https://doi.org/10.1007/BF01257083}

\bibitem{ftrain3}
Berk, R., Heidari, H., Jabbari, S., Joseph, M., Kearns, M.J., Morgenstern, J.,
  Neel, S., Roth, A.: A convex framework for fair regression. CoRR
  \textbf{abs/1706.02409} (2017), \url{http://arxiv.org/abs/1706.02409}

\bibitem{selfdriving}
Bojarski, M., Testa, D.D., Dworakowski, D., Firner, B., Flepp, B., Goyal, P.,
  Jackel, L.D., Monfort, M., Muller, U., Zhang, J., Zhang, X., Zhao, J., Zieba,
  K.: End to end learning for self-driving cars. CoRR  (2016),
  \url{http://arxiv.org/abs/1604.07316}

\bibitem{embedding}
Bolukbasi, T., Chang, K., Zou, J.Y., Saligrama, V., Kalai, A.T.: Man is to
  computer programmer as woman is to homemaker? debiasing word embeddings. In:
  Advances in Neural Information Processing Systems 29: Annual Conference on
  Neural Information Processing Systems 2016 (NeurIPS 2016), Barcelona, Spain.
  pp. 4349--4357 (2016),
  \url{http://papers.nips.cc/paper/6228-man-is-to-computer-programmer-as-woman-is-to-homemaker-debiasing-word-embeddings}

\bibitem{bound20}
Bunel, R., Lu, J., Turkaslan, I., Torr, P.H.S., Kohli, P., Kumar, M.P.: Branch
  and bound for piecewise linear neural network verification. J. Mach. Learn.
  Res.  \textbf{21},  42:1--42:39 (2020),
  \url{http://jmlr.org/papers/v21/19-468.html}

\bibitem{ftrain1}
Cava, W.L., Moore, J.: Genetic programming approaches to learning fair
  classifiers. Proceedings of the 2020 Genetic and Evolutionary Computation
  Conference  (2020)

\bibitem{bank}
Dua, D., Graff, C.: Bank marketing dataset at {UCI} machine learning repository
  (2017), \url{https://archive.ics.uci.edu/ml/datasets/Bank+Marketing}

\bibitem{census}
Dua, D., Graff, C.: Census income dataset at {UCI} machine learning repository
  (2017), \url{https://archive.ics.uci.edu/ml/datasets/adult}

\bibitem{credit}
Dua, D., Graff, C.: German credit dataset at {UCI} machine learning repository
  (2017),
  \url{https://archive.ics.uci.edu/ml/datasets/statlog+(german+credit+data)}

\bibitem{exactness19}
Dvijotham, K.D., Stanforth, R., Gowal, S., Qin, C., De, S., Kohli, P.:
  Efficient neural network verification with exactness characterization. In:
  Globerson, A., Silva, R. (eds.) Proceedings of the Thirty-Fifth Conference on
  Uncertainty in Artificial Intelligence, {UAI} 2019, Tel Aviv, Israel, July
  22-25, 2019. Proceedings of Machine Learning Research, vol.~115, pp.
  497--507. {AUAI} Press (2019),
  \url{http://proceedings.mlr.press/v115/dvijotham20a.html}

\bibitem{fairness}
Dwork, C., Hardt, M., Pitassi, T., Reingold, O., Zemel, R.S.: Fairness through
  awareness. In: Innovations in Theoretical Computer Science 2012, Cambridge,
  MA, USA. pp. 214--226 (2012), \url{https://doi.org/10.1145/2090236.2090255}

\bibitem{Ehlers17}
Ehlers, R.: Formal verification of piece-wise linear feed-forward neural
  networks. CoRR  \textbf{abs/1705.01320} (2017),
  \url{http://arxiv.org/abs/1705.01320}

\bibitem{abstract20}
Elboher, Y.Y., Gottschlich, J., Katz, G.: An abstraction-based framework for
  neural network verification. In: Lahiri, S.K., Wang, C. (eds.) Computer Aided
  Verification - 32nd International Conference, {CAV} 2020, Los Angeles, CA,
  USA, July 21-24, 2020, Proceedings, Part {I}. Lecture Notes in Computer
  Science, vol. 12224, pp. 43--65. Springer (2020).
  \doi{10.1007/978-3-030-53288-8\_3},
  \url{https://doi.org/10.1007/978-3-030-53288-8\_3}

\bibitem{demographic}
Feldman, M., Friedler, S.A., Moeller, J., Scheidegger, C., Venkatasubramanian,
  S.: Certifying and removing disparate impact. In: Proceedings of the 21th
  {ACM} {SIGKDD} International Conference on Knowledge Discovery and Data
  Mining, Sydney, NSW, Australia. pp. 259--268 (2015),
  \url{https://doi.org/10.1145/2783258.2783311}

\bibitem{fraud_detection}
Fu, K., Cheng, D., Tu, Y., Zhang, L.: Credit card fraud detection using
  convolutional neural networks. In: Neural Information Processing - 23rd
  International Conference ({ICONIP} 2016), Kyoto, Japan. pp. 483--490 (2016),
  \url{https://doi.org/10.1007/978-3-319-46675-0\_53}

\bibitem{goodt}
Gale, W.: Good-turing smoothing without tears. Journal of Quantitative
  Linguistics pp. 217--37 (1995)

\bibitem{THEMIS}
Galhotra, S., Brun, Y., Meliou, A.: Fairness testing: Testing software for
  discrimination. In: Proceedings of the 2017 11th Joint Meeting on Foundations
  of Software Engineering ({ESEC/FSE} 2017), Paderborn, Germany. pp. 498--510
  (2017), \url{https://doi.org/10.1145/3106237.3106277}

\bibitem{counterfactual}
Garg, S., Perot, V., Limtiaco, N., Taly, A., Chi, E.H., Beutel, A.:
  Counterfactual fairness in text classification through robustness. In:
  Proceedings of the 2019 {AAAI/ACM} Conference on AI, Ethics, and Society
  ({AIES} 2019), Honolulu, HI, USA. pp. 219--226 (2019),
  \url{https://doi.org/10.1145/3306618.3317950}

\bibitem{ai2}
Gehr, T., Mirman, M., Drachsler{-}Cohen, D., Tsankov, P., Chaudhuri, S.,
  Vechev, M.T.: {AI2:} safety and robustness certification of neural networks
  with abstract interpretation. In: 2018 {IEEE} Symposium on Security and
  Privacy, {SP} 2018, Proceedings, 21-23 May 2018, San Francisco, California,
  {USA}. pp. 3--18. {IEEE} Computer Society (2018).
  \doi{10.1109/SP.2018.00058}, \url{https://doi.org/10.1109/SP.2018.00058}

\bibitem{minimal20}
Goldberger, B., Katz, G., Adi, Y., Keshet, J.: Minimal modifications of deep
  neural networks using verification. In: Albert, E., Kov{\'{a}}cs, L. (eds.)
  {LPAR} 2020: 23rd International Conference on Logic for Programming,
  Artificial Intelligence and Reasoning, Alicante, Spain, May 22-27, 2020. EPiC
  Series in Computing, vol.~73, pp. 260--278. EasyChair (2020),
  \url{https://easychair.org/publications/paper/CWhF}

\bibitem{ensemble20}
Gross, D., Jansen, N., P{\'{e}}rez, G.A., Raaijmakers, S.: Robustness
  verification for classifier ensembles. In: Hung, D.V., Sokolsky, O. (eds.)
  Automated Technology for Verification and Analysis - 18th International
  Symposium, {ATVA} 2020, Hanoi, Vietnam, October 19-23, 2020, Proceedings.
  Lecture Notes in Computer Science, vol. 12302, pp. 271--287. Springer (2020).
  \doi{10.1007/978-3-030-59152-6\_15},
  \url{https://doi.org/10.1007/978-3-030-59152-6\_15}

\bibitem{invariant20}
Jacoby, Y., Barrett, C.W., Katz, G.: Verifying recurrent neural networks using
  invariant inference. In: Hung, D.V., Sokolsky, O. (eds.) Automated Technology
  for Verification and Analysis - 18th International Symposium, {ATVA} 2020,
  Hanoi, Vietnam, October 19-23, 2020, Proceedings. Lecture Notes in Computer
  Science, vol. 12302, pp. 57--74. Springer (2020).
  \doi{10.1007/978-3-030-59152-6\_3},
  \url{https://doi.org/10.1007/978-3-030-59152-6\_3}

\bibitem{synonym_2}
Jia, R., Liang, P.: Adversarial examples for evaluating reading comprehension
  systems. In: Proceedings of the 2017 Conference on Empirical Methods in
  Natural Language Processing ({EMNLP} 2017), Copenhagen, Denmark. pp.
  2021--2031 (2017), \url{https://doi.org/10.18653/v1/d17-1215}

\bibitem{Reluplex}
Katz, G., Barrett, C.W., Dill, D.L., Julian, K., Kochenderfer, M.J.: Reluplex:
  An efficient {SMT} solver for verifying deep neural networks. In: Majumdar,
  R., Kuncak, V. (eds.) Computer Aided Verification - 29th International
  Conference, {CAV} 2017, Heidelberg, Germany, July 24-28, 2017, Proceedings,
  Part {I}. Lecture Notes in Computer Science, vol. 10426, pp. 97--117.
  Springer (2017). \doi{10.1007/978-3-319-63387-9\_5},
  \url{https://doi.org/10.1007/978-3-319-63387-9\_5}

\bibitem{marabou20}
Katz, G., Huang, D.A., Ibeling, D., Julian, K., Lazarus, C., Lim, R., Shah, P.,
  Thakoor, S., Wu, H., Zeljic, A., Dill, D.L., Kochenderfer, M.J., Barrett,
  C.W.: The marabou framework for verification and analysis of deep neural
  networks. In: Dillig, I., Tasiran, S. (eds.) Computer Aided Verification -
  31st International Conference, {CAV} 2019, New York City, NY, USA, July
  15-18, 2019, Proceedings, Part {I}. Lecture Notes in Computer Science, vol.
  11561, pp. 443--452. Springer (2019). \doi{10.1007/978-3-030-25540-4\_26},
  \url{https://doi.org/10.1007/978-3-030-25540-4\_26}

\bibitem{ftrain4}
Kearns, M.J., Neel, S., Roth, A., Wu, Z.S.: Preventing fairness gerrymandering:
  Auditing and learning for subgroup fairness. In: Dy, J.G., Krause, A. (eds.)
  Proceedings of the 35th International Conference on Machine Learning, {ICML}
  2018, Stockholmsm{\"{a}}ssan, Stockholm, Sweden, July 10-15, 2018.
  Proceedings of Machine Learning Research, vol.~80, pp. 2569--2577. {PMLR}
  (2018), \url{http://proceedings.mlr.press/v80/kearns18a.html}

\bibitem{pso1995}
{Kennedy}, J., {Eberhart}, R.: Particle swarm optimization. In: Proceedings of
  ICNN'95 - International Conference on Neural Networks. vol.~4, pp. 1942--1948
  vol.4 (1995). \doi{10.1109/ICNN.1995.488968}

\bibitem{popqorn}
Ko, C., Lyu, Z., Weng, L., Daniel, L., Wong, N., Lin, D.: {POPQORN:}
  quantifying robustness of recurrent neural networks. In: Chaudhuri, K.,
  Salakhutdinov, R. (eds.) Proceedings of the 36th International Conference on
  Machine Learning, {ICML} 2019, 9-15 June 2019, Long Beach, California, {USA}.
  Proceedings of Machine Learning Research, vol.~97, pp. 3468--3477. {PMLR}
  (2019), \url{http://proceedings.mlr.press/v97/ko19a.html}

\bibitem{pmc}
Kwiatkowska, M., Norman, G., Parker, D.: Advances and challenges of
  probabilistic model checking. 2010 48th Annual Allerton Conference on
  Communication, Control, and Computing, Allerton 2010  (09 2010).
  \doi{10.1109/ALLERTON.2010.5707120}

\bibitem{smc}
Legay, A., Delahaye, B., Bensalem, S.: Statistical model checking: An overview.
  In: Barringer, H., Falcone, Y., Finkbeiner, B., Havelund, K., Lee, I., Pace,
  G., Ro{\c{s}}u, G., Sokolsky, O., Tillmann, N. (eds.) Runtime Verification.
  pp. 122--135. Springer Berlin Heidelberg, Berlin, Heidelberg (2010)

\bibitem{textbugger}
Li, J., Ji, S., Du, T., Li, B., Wang, T.: Textbugger: Generating adversarial
  text against real-world applications. In: 26th Annual Network and Distributed
  System Security Symposium ({NDSS} 2019), San Diego, California, USA (2019),
  \url{https://www.ndss-symposium.org/ndss-paper/textbugger-generating-adversarial-text-against-real-world-applications/}

\bibitem{kmeans}
Lloyd, S.P.: Least squares quantization in {PCM}. {IEEE} Trans. Information
  Theory  \textbf{28}(2),  129--136 (1982),
  \url{https://doi.org/10.1109/TIT.1982.1056489}

\bibitem{metamorphic}
Ma, P., Wang, S., Liu, J.: Metamorphic testing and certified mitigation of
  fairness violations in {NLP} models. In: Bessiere, C. (ed.) Proceedings of
  the Twenty-Ninth International Joint Conference on Artificial Intelligence,
  {IJCAI} 2020. pp. 458--465. ijcai.org (2020). \doi{10.24963/ijcai.2020/64},
  \url{https://doi.org/10.24963/ijcai.2020/64}

\bibitem{glove}
Pennington, J., Socher, R., Manning, C.D.: Glove: Global vectors for word
  representation. In: Proceedings of the 2014 Conference on Empirical Methods
  in Natural Language Processing ({EMNLP} 2014), October 25-29, 2014, Doha,
  Qatar. pp. 1532--1543 (2014),
  \url{https://www.aclweb.org/anthology/D14-1162/}

\bibitem{socrates}
Pham, L.H., Li, J., Sun, J.: {SOCRATES:} towards a unified platform for neural
  network verification. CoRR  \textbf{abs/2007.11206} (2020),
  \url{https://arxiv.org/abs/2007.11206}

\bibitem{gensim}
{\v R}eh{\r u}{\v r}ek, R., Sojka, P.: {Software Framework for Topic Modelling
  with Large Corpora}. In: {Proceedings of the LREC 2010 Workshop on New
  Challenges for NLP Frameworks}, Valletta, Malta. pp. 45--50 (2010),
  \url{http://is.muni.cz/publication/884893/en}

\bibitem{face_recognition}
Schroff, F., Kalenichenko, D., Philbin, J.: Facenet: {A} unified embedding for
  face recognition and clustering. In: {IEEE} Conference on Computer Vision and
  Pattern Recognition ({CVPR} 2015), Boston, MA, USA. pp. 815--823 (2015),
  \url{https://doi.org/10.1109/CVPR.2015.7298682}

\bibitem{pso1998}
Shi, Y., Eberhart, R.: Parameter selection in particle swarm optimization. In:
  Evolutionary Programming (1998)

\bibitem{deeppoly}
Singh, G., Gehr, T., P{\"{u}}schel, M., Vechev, M.T.: An abstract domain for
  certifying neural networks. Proc. {ACM} Program. Lang.  \textbf{3}({POPL}),
  41:1--41:30 (2019). \doi{10.1145/3290354},
  \url{https://doi.org/10.1145/3290354}

\bibitem{patch19}
Sotoudeh, M., Thakur, A.: Correcting deep neural networks with small,
  generalizing patches. In: Workshop on Safety and Robustness in Decision
  Making (2019)

\bibitem{science}
Thomas, P.S., da~Silva, B.C., Barto, A.G., Giguere, S., Brun, Y., Brunskill,
  E.: Preventing undesirable behavior of intelligent machines. Science
  \textbf{366}(6468),  999--1004 (2019),
  \url{https://science.sciencemag.org/content/366/6468/999}

\bibitem{milp}
Tjeng, V., Xiao, K.Y., Tedrake, R.: Evaluating robustness of neural networks
  with mixed integer programming. In: 7th International Conference on Learning
  Representations, {ICLR} 2019, New Orleans, LA, USA, May 6-9, 2019.
  OpenReview.net (2019), \url{https://openreview.net/forum?id=HyGIdiRqtm}

\bibitem{fairtest}
Tram{\`{e}}r, F., Atlidakis, V., Geambasu, R., Hsu, D.J., Hubaux, J., Humbert,
  M., Juels, A., Lin, H.: Fairtest: Discovering unwarranted associations in
  data-driven applications. In: 2017 {IEEE} European Symposium on Security and
  Privacy (EuroS{\&}P 2017), Paris, France. pp. 401--416 (2017),
  \url{https://doi.org/10.1109/EuroSP.2017.29}

\bibitem{aequitas2018}
Udeshi, S., Arora, P., Chattopadhyay, S.: Automated directed fairness testing.
  In: Huchard, M., K{\"{a}}stner, C., Fraser, G. (eds.) Proceedings of the 33rd
  {ACM/IEEE} International Conference on Automated Software Engineering, {ASE}
  2018, Montpellier, France, September 3-7, 2018. pp. 98--108. {ACM} (2018).
  \doi{10.1145/3238147.3238165}, \url{https://doi.org/10.1145/3238147.3238165}

\bibitem{FairerML}
Veale, M., Binns, R.: Fairer machine learning in the real world: Mitigating
  discrimination without collecting sensitive data. Big Data \& Society
  \textbf{4} (2017)

\bibitem{medical_diagnosis}
Vieira, S., Pinaya, W.H., Mechelli, A.: Using deep learning to investigate the
  neuroimaging correlates of psychiatric and neurological disorders: Methods
  and applications. Neuroscience \& Biobehavioral Reviews  \textbf{74},  58--75
  (2017), \url{https://doi.org/10.1016/j.neubiorev.2017.01.002}

\bibitem{symbolic18}
Wang, S., Pei, K., Whitehouse, J., Yang, J., Jana, S.: Formal security analysis
  of neural networks using symbolic intervals. In: Enck, W., Felt, A.P. (eds.)
  27th {USENIX} Security Symposium, {USENIX} Security 2018, Baltimore, MD, USA,
  August 15-17, 2018. pp. 1599--1614. {USENIX} Association (2018),
  \url{https://www.usenix.org/conference/usenixsecurity18/presentation/wang-shiqi}

\bibitem{adf2020}
Zhang, P., Wang, J., Sun, J., Dong, G., Wang, X., Wang, X., Dong, J.S., Dai,
  T.: White-box fairness testing through adversarial sampling. Proceedings of
  the 42th International Conference on Software Engineering ({ICSE} 2020),
  Seoul, South Korea  (2020)

\end{thebibliography}

\bibliographystyle{splncs04}

\appendix
\section{Appendix}
\subsection{Preliminary}
\noindent \emph{Neural Networks} In general, a neural network can be viewed as a function $N: \mathbb{R}^{p} \to \mathbb{R}^{q}$ which maps an input $x\in \mathbb{R}^{p}$ (e.g., images or texts) to an output $y \in \mathbb{R}^{q}$ (e.g., predictions for image classification or texts for machine translation). In this work, we focus on deep feed-forward neural networks and recurrent neural networks. We leave other neural network models to the future work. 

Neural networks usually follow a layered architecture, where the computational nodes, a.k.a.~neurons, are organized layer-wise and data flows from layer to layer. The first layer is the input layer; the last layer is the output layer and the remaining are hidden layers. In the case of recurrent neural networks, the output of a hidden neuron may be connected to the input of a neuron of a previous layer, i.e., forming a feedback loop. 

Based on the transformation that a layer performs, there are two commonly used layers: affine layers and activation layers. An affine layer applies an affine transformation i.e., $\pi(x) = Wx + b$ where $x$ is the input from the previous layer; $W$ is a weight matrix and $b$ is a bias. An activation layer applies a non-linear activation function $\sigma$. Commonly applied activation functions include Rectified Linear Unit (ReLU) $\sigma(x) = max(0, x)$, Sigmoid $\sigma(x) = \frac{e^x}{e^x+1}$ and Tanh $\sigma(x) = \frac{e^x-e^{-x}}{e^x+e^{-x}}$. These functions are applied neuron-wise, e.g., given an input $x = (x_0, \cdots , x_{p-1}) \in \mathbb{R}^{p}$, $\sigma(x) = \big(\sigma(x_0), \cdots , \sigma(x_{p-1})\big)$. In the following, we assume a neural network $N$ consists of $n$ layers, and each layer $i$ contains $d_i$ neurons. Then layer $i$ is a function $f_i: \mathbb{R}^{d_{i}} \to \mathbb{R}^{d_{i+1}}$ mapping the input of layer $i$, i.e., $x_i$, to the input of layer $i+1$, i.e., $x_{i+1}$, and the neural network is function $N: \mathbb{R}^{d_0} \to \mathbb{R}^{d_{n}}$, where $d_{n}$ is the dimension of output. Usually the input layer does not transform the data, and thus $f_0=I$. \\

\subsection{Definition of DTMC}\label{A:DTMC}
\begin{definition}[Discrete-Time Markov Chain (DTMC)]
\label{def:dtmc}
A Discrete-Time Markov Chain is a tuple $M=(S, I, A)$, where:
\begin{itemize}
\item $S$ is a finite set of states;
\item $I: S \rightarrow [ 0, 1 ]$ is the initial distribution: $ \sum_{s \in S} I(s) = 1$;
\item $A: S \times S \rightarrow [ 0, 1 ]$ is the transition probability matrix and for every state $s \in S$, $\sum_{s' \in S} A(s, s') = 1$.
\end{itemize}
\end{definition}

\subsection{Proof of Theorem~\ref{main} }\label{A:proof}
\begin{proof}
By Theorem 6 in~\cite{DBLP:conf/cav/BazilleGJS20}, for all $p \in S, \; n_p \geq (\frac{11}{10}B(A_W^\alpha))^2H(n)$, we have for any CTL property $\phi$:

\begin{equation}
P(\big| \gamma(A, \phi) - \gamma(A_W, \phi)\big | > \epsilon) \leq \delta
\end{equation}

Unlike~\cite{DBLP:conf/cav/BazilleGJS20} which uses Laplace smoothing to learn the model, we adopted the above-defined estimation in this work. Hence, conditioning $Cond_s^l(A)$ and Laplace offset which are used to take into account the error introduced by Laplace smoothing is not necessary in our approach. Hence, we can omit the part $(\frac{11}{10}B(A_W^\alpha))^2$ and set it to 1 instead. Thus, by satisfying the stopping criteria  $n_p \geq H(n)$, we have for any CTL property $\psi$ that $P(\big| \gamma(A, \psi) - \gamma(A_W, \psi)\big | > \epsilon) \leq \delta$. 
\hfill \qed
\end{proof} 

\subsection{Proof of Theorem~\ref{the:probdiff}}\label{A:proofprobdiff}
\begin{proof}

Since 
$P(|X - X_t| > \epsilon) \leq \delta$ and $P(|Z - Z_t| > \epsilon) \leq \delta$, we have
\begin{center}
$P(|X - X_t| \leq \epsilon) \geq 1 - \delta$\\
$P(|Z - Z_t| \leq \epsilon) \geq 1 - \delta$
\end{center}
Hence
\begin{center}
$P(|(X - X_t) - (Z - Z_t)| \leq 2\epsilon) \geq P(|X - X_t|\leq \epsilon) \cdot P(|Z - Z_t|\leq \epsilon) \geq (1-\delta)^2$
\end{center}
and
\begin{center}
$P(|(X - X_t) - (Z - Z_t)| > 2\epsilon) \leq 1 - (1-\delta)^2$
\end{center}
Finally
\begin{equation*}
P(|(X - Z)-(X_t - Z_t)| > 2\epsilon) \leq 2\delta - \delta^2 
\end{equation*}
\hfill \qed
\end{proof}

\subsection{Proof of Theorem~\ref{the:pac}}\label{A:proof3}
\begin{proof}
By Theorem~\ref{main}, we have PAC-correct probability $P(Y=l\: | \: F=f_i), \forall \; f_i \in F$ with accuracy $\epsilon = \frac{\mu\epsilon}{2}$ and confidence $\delta = 1-\sqrt{1-\mu\delta}$. Next, by Theorem \ref{the:probdiff}, we have probability difference between any pair of $f_i$ is PAC-correct with accuracy $2\epsilon$ and confidence $2\delta - \delta^2$. Hence Algorithm 1 is PAC-correct with accuracy $\mu\epsilon$ and confidence $\mu\delta$. \hfill \qed
\end{proof}

\subsection{Example Sensitivity Analysis Result}\label{A.eg}
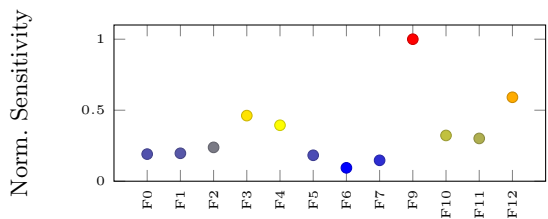
\begin{figure}[t]
                    \begin{tikzpicture}
                    \pgfplotsset{%
                        width=0.6\textwidth,
                        height=0.3\textwidth
                    }
                    \pgfplotsset{every tick label/.append style={font=\tiny}}
                    \begin{axis}[
                        ylabel={Norm. Sensitivity},
                        xmin=0, xmax=13,
                        ymin=0, ymax=1.1,
                        xtick = {1,2,3,4,5,6,7,8,9,10,11,12},
                        xticklabels={F0,F1,F2,F3,F4,F5,F6,F7,F9,F10,F11,F12,},
                        x tick label style={rotate=90,anchor=east}],
                        ytick={0,0.02,0.04,0.06,0.08},
                        legend pos=north west,
                        ymajorgrids=false,
                        grid style=dashed,
                    ]
                    
                    \addplot[
                        only marks,
                        scatter,    
                        ]
                        coordinates {
                        (1,0.1909)(2,0.1967)(3,0.2383)(4,0.4620)(5,0.3941)(6,0.1827)(7,0.0943)(8,0.1471)(9,1)(10,0.3223)(11,0.3013)(12,0.5911)
                        };
                        
                    \end{axis}
                    \end{tikzpicture}
            \centering 
            \caption{Example sensitivity analysis result}
            \label{fig:seneg}
\end{figure}

Figure~\ref{fig:seneg} shows the sensitivity analysis result on non-protected features for our running example.

\subsection{Particle Swarm Optimization}\label{A:PSO}
In PSO, multiple particles are placed in the search space. At each time step, each particle updates its location $\overrightarrow{x_i}$ and velocity $\overrightarrow{v_i}$ according to an objective function. That is, the velocity is updated based on the current velocity $\overrightarrow{v_i}$, the previous best location found locally $\overrightarrow{p_i}$ and the previous best location found globally $\overrightarrow{p_g}$. Its location is updated based on the current location and velocity. We write $R(0,c)$ to denote a random number uniformly sampled from the range of $[0, c]$. Formally, the PSO update equation is as follows~\cite{pso1998}.
\begin{align}
    \overrightarrow{v_i} & \leftarrow \omega\overrightarrow{v_i} + R(0, c_1)(\overrightarrow{p_i} - \overrightarrow{x_i}) + R(0, c_2)(\overrightarrow{p_g} - \overrightarrow{x_i}) \\
    \overrightarrow{x_i} & \leftarrow \overrightarrow{x_i} + \overrightarrow{v_i}
\end{align}
where $\omega$, $c_1$, $c_2$ as inertia weight, cognitive parameter and social parameter respectively.

\subsection{Thread to validity}
In our experiments, only 4 datasets are applied to evaluate the effectiveness of our approach. Although they are commonly used datasets for machine learning fairness studies, it may not be safe to generalize the conclusion to other models and datasets. This issue can be addressed with more models and datasets as well as protected features. Furthermore, in the experiments, we focus on FFNN and RNN only. Our approach can be potentially adopted for other deep learning models, such as CNN. This is because our approach only requires a way of identifying a set of abstract states and sampling traces. Lastly, our approach requires white-box access to the model to perform analysis and repairing. Hence our approach is not applicable to blackbox models. 

\end{document}